\documentclass{article} 
\usepackage{geometry}
\usepackage[english]{babel} 
\usepackage{amssymb}
\usepackage{amsmath}
\usepackage{multirow, makecell}
\usepackage{bigstrut}
\usepackage{txfonts}
\usepackage{subfigure}
\usepackage{multicol}
\usepackage{booktabs}
\usepackage{mathdots}
\usepackage{graphicx} 
\usepackage{fancyhdr}
\usepackage{hyperref}
\usepackage{float}
\usepackage{microtype}
\usepackage{caption}
\usepackage[numbers]{natbib}
\usepackage[ruled,vlined]{algorithm2e}

\makeatletter
\def\@xfootnote[#1]{%
  \protected@xdef\@thefnmark{#1}%
  \@footnotemark\@footnotetext}
\makeatother

\usepackage{sectsty}
\subsectionfont{\normalfont\itshape}
\subsubsectionfont{\normalfont\itshape}

\begin{document}


\begin{tabular}{p{1.1in}p{4.5in}p{1.2in}}  
\hspace{-1cm}
\noindent
\begin{tabular}{c}  \includegraphics[width=2.9cm]{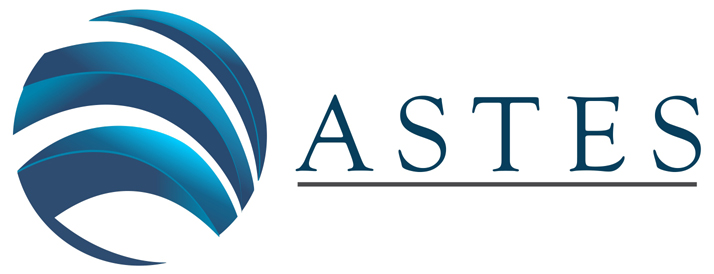}\end{tabular} 	& \vspace{-0.5cm} \centering \textit{Advances in Science, Technology and Engineering Systems Journal \newline Vol. 5, No. 5, XX-YY (2020)} \\   \href{http://www.astesj.com}{www.astesj.com}  
	& \vspace{-0.6cm}  \rule{1.2in}{1.5pt} \vspace{-0.2cm} \newline \centering \textbf{ ASTES Journal \newline ISSN: 2415-6698} \newline \rule{1.2in}{1.7pt} 
\end{tabular}

\vspace{1.8cm}

\noindent \textbf{\LARGE{\setlength\itemsep{0pt}Dilated Fully Convolutional Neural Network for Depth Estimation from a Single Image}}

\vspace{0.2cm}

\noindent Binghan Li\footnote[*]{Binghan Li, Department of Electrical \& Computer Engineering, Texas A\&M University, College Station, TX, 77840, USA, Email: libinghan1994@outlook.com, }${}^{,1}$, Yindong Hua${}^{2}$, Yifeng Liu${}^{3}$, Mi Lu${}^{4}$   

\vspace{0.2cm}
\noindent\textit{${}^{1}$Department of Electrical \& Computer Engineering, Texas A\&M University, College Station, TX, 77840, USA}

\vspace{0.2cm}
\noindent\textit{${}^{2}$Department of Electrical \& Computer Engineering, Stony Brook University, Stony Brook, NY, 11794, USA}

\vspace{0.2cm}
\noindent\textit{${}^{3}$School of Business, Stevens Institute of Technology, Hoboken, NJ, 07030, USA}

\vspace{0.2cm}
\noindent\textit{${}^{4}$Department of Electrical \& Computer Engineering, Texas A\&M University, College Station, TX, 77840, USA}

\vspace{0.3cm}

\noindent\begin{tabular}{p{1.7in} p{0.1in} p{5.2in} }
A R T I C L E \hspace{0.1cm} I N F O &  & A B S T R A C T \\ 
 \cline{1-1}  \cline{3-3} \setlength\itemsep{0pt} \vspace{-0.1cm}
\textit{Article history:
	\newline Received: 25 December, 2020
	\newline Accepted: 04 March, 2021
	\newline Online: 12 March, 2021
	\newline \rule{1.78in}{0.5pt} 
	Keywords: 
	\newline Depth Prediction
	\newline CNN
	\newline Dilated Convolutions}
 \newline \newline  & & \vspace{-0.1cm} 
 \textit{Depth prediction plays a key role in understanding a 3D scene. Several techniques have been developed throughout the years, among which Convolutional Neural Network has recently achieved state-of-the-art performance on estimating depth from a single image. However, traditional CNNs suffer from the lower resolution and information loss caused by the pooling layers. And oversized parameters generated from fully connected layers often 
lead to a exploded memory usage problem. In this paper, we present an advanced Dilated Fully Convolutional Neural Network to address the deficiencies. Taking advantages of the exponential expansion of the receptive field in dilated convolutions, our model can minimize the loss of resolution. It also reduces the amount of parameters significantly by replacing the fully connected layers with the fully convolutional layers. We show experimentally on NYU Depth V2 datasets that the depth prediction obtained from our model is considerably closer to ground truth than that from traditional CNNs techniques.}\\
 \cline{1-1}  \cline{3-3}
\end{tabular}

\vspace{0.5cm}

\begin{multicols}{2}

\section{Introduction}
This paper is an extension of work originally presented in conference name \cite{hua2019dilated}. Depth prediction has always been a core task to understand the geometric relations within a 3D scene. It provides rich information about the distance of the objects in the image from the viewpoint of camera. This technique is necessary for many applications in computer vision including smoothing blurred parts of an image \cite{he2010single} \cite{li2018multiple}, rendering of 3D scenes \cite{saxena2009make3d}, virtual reality, self-driving cars \cite{hadsell2009learning}, grasping in robotics \cite{ye2017self} and autopilot \cite{tateno2017cnn}. However, predicting depth from images is a quite complex and challenging task. In absence of the environmental assumptions, the inherent ambiguity of mapping an intensity or color measurement into a depth value makes depth prediction an ill-posed problem. Many unique techniques have been proposed to tackle this problem, such as superpixels based algorithms \cite{liu2014discrete}, Structure-from-Motion (SfM) \cite{szeliski2011structure}, data-driven methods \cite{karsch2012depth} and CNN based approaches \cite{laina2016deeper}. 

Neural Network has been widely applied on computer vision tasks and natural language processing tasks \cite{wang2017combining} \cite{wang2017comparisons} \cite{wang2020attention}. And Convolutional Neural Network (CNN) has achieved a great success on outperforming many state-of-the-art algorithms over object classification and detection \cite{he2017mask}, semantic segmentation \cite{chen2014semantic}, scene reconstruction, and face recognition \cite{jiang2017face}. Recently, depth prediction can also be addressed by CNNs due to the power that the ambiguous mapping between a single image and depth maps can be modeled via learning in the neural network. \cite{laina2016deeper} increases the output resolution by efficiently learning the feature map up-sampling within the fully convolutional residual networks. The Multiple-Scale Deep Neural Network in \cite{eigen2014depth} improves the depth prediction by estimating the global scene structure and refining it using local information. Three different computer vision tasks including depth prediction are addressed in \cite{eigen2015predicting} with a single multi-scale convolutional network architecture. \cite{godard2017unsupervised} proposes a novel training loss in the CNN architecture, which enforces consistency between left and right depth maps, to perform end-to-end unsupervised single image depth estimation.

Despite Convolutional Neural Networks performs well on depth estimation, there are still some deficiencies remained to be improved. Some depth predicting CNN models still use pooling layers to extend receptive field. Pooling layers provide an effective approach to reduce the dimensionality of the network by summarizing the presence of features in patches of the feature map. However, they inevitably lose a lot of valuable information during the down sampling process. In contrast, fully connected layers will obtain the global relationship between pixels and image, and inherit all the combinations of the features from the previous layer. This will generate too many parameters, making fully connected layers incredibly computationally expensive. 

\begin{figure}[H]
	\subfigure[1-dilated convolution]
	{
		\includegraphics[width=1.6in]{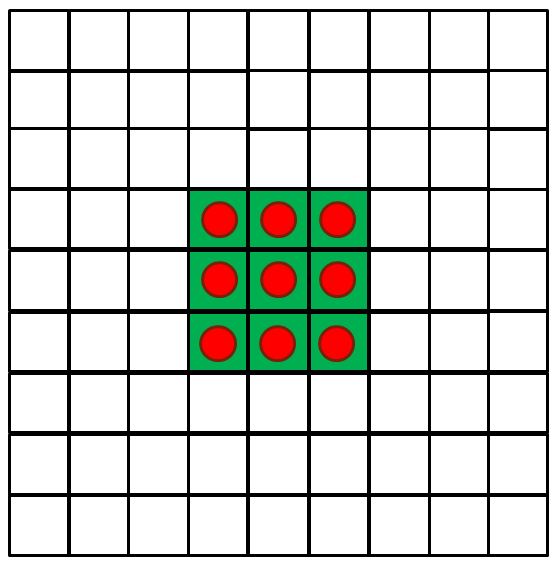}
		\label{b_1}
	}
	\subfigure[2-dilated convolution]
	{
		\includegraphics[width=1.6in]{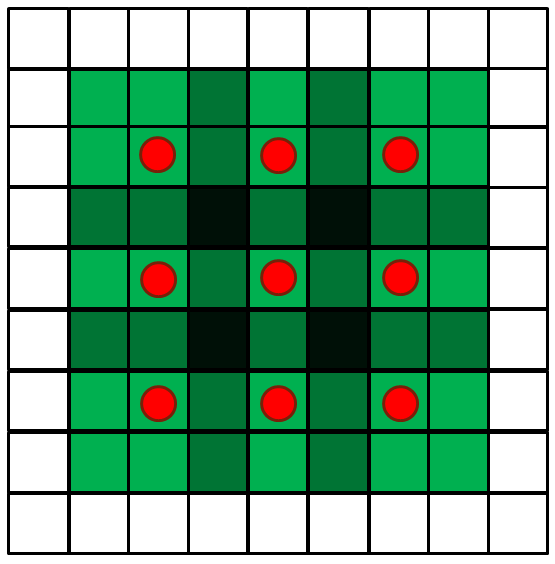}
		\label{b_2}
	}
	\caption{Examples of dilated convolutions with receptive field. (a) refers to an 1-dilated convolution with a kernel of a 3x3 receptive field, which is inherently a standard convolution. (b) refers to a 2-dilated convolutions with a kernel of a 7x7 receptive field. The amount of parameters corresponding to each layer stays identical. }
	\label{fig:The feature of receptive field in the dilated convolutions }
\end{figure}

Dilated convolutional neural network has been proved to be more effective than traditional CNNs in \cite{chen2014semantic} \cite{yu2015multi}.The feature of dilated convolutions is illustrated in Figure \ref{fig:The feature of receptive field in the dilated convolutions }. Instead of contiguous pooling filters in traditional CNNs, dilation imposes filters that have spaces between each node. The dilated convolutions support exponential expansion of the receptive field without loss of resolution or coverage, which reduces the computation and memory costs. While preserving the dimensions of data at the output layer, dilated convolutions also maintain the ordering of data. 

In this paper, a fully dilated convolutional neural network is proposed to enhance the performance of depth prediction. Our model is designed based on the framework in \cite{eigen2014depth}. It contains two components: the global coarse-scale network, which estimates the coarse depth, and the local fine-scale network, which refines the coarse depth estimation combined with local information. In the coarse-scale network, the convolutional layers are replaced with the dilated convolutional layers and the fully connected layers are replaced with the convolutional layers. We train and evaluate our network on NYU Depth V2 datasets and make comparison on the depth predicting performances between our model with traditional VGG-16 model. The experiment results demonstrate that our model reduces the computational and memory costs, and achieves state-of-the-art performance on benchmarks.  

\section{Related Work}

\subsection{Depth Estimation}
Depth estimation refers to a set of techniques and algorithms aiming to obtain the spatial information from a scene, which has been an essential computer vision task with a long history. Normally a 2D image taking from the camera is not able to represent the local spatial relations of a 3D scene. Since only one point of each pixel is projected in the real scene, its depth information is mathematically eliminated when projected into plane in a image. Depth information is a key prerequisite to perform multiple computer vision tasks.  With the depth information, we are able to back project images captured from multiple views and the 3D scenes can be perfectly restructured by matching all the points. In order to accurately move the actuators in robotics, the depth estimation is required to multiple tasks such as perception, navigation, and planning. In the single image dehazing task, scene depth is a key parameter that supports to remove the haze locally instead of globally. 

Depth estimation is an inherently ambiguous and complex task. Geometrically, infinite points in the scene are not projected, thus the depth information may be generated from considerable possible world scenes. For some specific computer vision tasks benefited from depth prediction, it will face more challenges. In the application of autonomous driving in particular, depth estimation will be degraded by occlusion, dynamic object in the scene and imperfect stereo correspondence. Despite the loss of depth information in the 3D dimension, depth map prediction has been investigated inspired by the analogy to how human eyes perceive depth information from depth cues. 

\subsection{State-of-the-art Algorithms}
Humans estimate depth by comparing the images obtained from left and right eyes. Our eyes perceive and recognize the depth cues of the scene, then our brains will subconsciously analyze and recover the depth information easily. There are basically 4 categories of depth cues: Static monocular, depth from motion, binocular and physiological cues \cite{kooi2004visual}, making it possible to construct the spatial arrangement of objects in the scene. In computer vision, depth information is extracted mainly from monocular images and stereo images by exploiting epipolar geometry. And many efforts have been made to tackle the challenge of predicting depth from  stereo vision depth estimation and monocular depth estimation. 

Stereo vision is inspired by how human eyes calculate the approximate depth map from the minor difference between both viewpoints. It compares two differing views on a scene and predicts the relative depth information from the displacement in horizontal coordinates of corresponding image points. The local disparity can be obtained effortlessly using local appearance features. By contrast, estimating depth information from a single image requires a global view of the scene, which is one of the reasons why the monocular depth estimation has not been solved to the same degree as the stereo vision approach \cite{eigen2014depth}. In \cite{scharstein2002taxonomy}, Scharstein summarizes a taxonomy and evaluation of existing dense two-frame stereo correspondence algorithms, including some earliest stereo vision based depth estimation algorithms \cite{boykov1998variable} \cite{okutomi1993multiple} with impressive performances. \cite{memisevic2011stereopsis} proposes a probabilistic deep learning approach to model disparity and generate binocular training data to estimate model parameters, which outperforms state-of-the-art algorithms with fewer requirements for global detailed information of the scene. 

Structure-from-Motion (SfM) is one of the most successful state-of-the-art techniques of depth estimation \cite{ozyesil2017survey}. SfM estimates the camera motion from the relative pairwise camera positions of the extracted features. Then it predicts the depth information via triangulation from pairs of consecutive views, and recovers the 3D structure through the spatial and geometric relationship of objects in the scene \cite{szeliski2011structure}. Despite its advantages over the lower cost and the less restricted environment, SfM has not been widely adopted in commercial applications due to its complex theories and the difficulty to further enhance the accuracy and speed. 

Several techniques address the depth estimation task based on superpixels. \cite{felzenszwalb2004efficient} proposes an algorithm to find an oversegmentation of the image that breaks up the planar surfaces into many small patches, which are named as superpixels. For each homogeneous superpixel in the image, \cite{saxena2009make3d} uses a Markov Random Field (MRF) to infer depth information through a set of plane coefficients, which extract both the 3D location and orientation in the image. The MRF is trained via supervised learning to learn how different depth cues are associated with different depths. \cite{saxena2009make3d} further extends the model by combining triangulation cues and monocular images cues, which supports to restructure a full and photorealistic 3D model of a larger scene. There are also some superpixels based algorithms deploying the Conditional Random Fields (CRFs) for the regularization of depth information. Deriving from the CRFs, \cite{liu2014discrete} formulates monocular depth estimation as a discrete-continuous optimization problem. The continuous variables encode the depth of the superpixels in the input image, and the discrete variables represent the relationships between neighboring superpixels. \cite{Li_2015_CVPR} extracts multi-scale image patches around the superpixel center and learn to encode the correlations between input patches and corresponding depths regressively with a deep CNN. Then it refines the depth estimation from the superpixel level to pixel level by using CRFs.

\subsection{CNN Based Algorithms}

A CNN-based depth estimation from a single image is a challenging task if without the local correspondences. A CNN model needs to self-learn the pixel-wise local details as well as the correlations between a pixel and the global scene during the training process.

\cite{eigen2014depth} presents a novel network to regress dense depth maps by implementing a deep network with two stages: global coarse-scale network and local fine-scale network. The coarse-scale network is initialized based on AlexNet \cite{krizhevsky2017imagenet} with five feature extraction layers, including convolution layers and max-pooling layers, and two fully connected layers followed. The network is able to integrate the understanding of a global view by making effective use of depth cues to predict the coarse depth information. The fine-scale network consists of convolutional layers and one pooling stage for the first layer edge features. Aligning with local details in the scene, the fine network fine-tune the coarse prediction by concatenating an additional low-level feature map.  

Building upon the two-scale CNN architecture in \cite{eigen2014depth}, \cite{eigen2015predicting} authors a paper about predicting depth, surface normals and semantic labels with an improved multi-scale convolutional network. It replaces the Alexnet network with a deeper VGG-16 network \cite{simonyan2014very}. Generating pixel-maps directly from an input image, this network can also align to many image details by using a sequence of convolutional network stacks applied at increasing resolution, without the need for low-level superpixels or contours. 

A novel fully convolutional network incorporated with efficient residual up-sampling blocks is proposed in \cite{laina2016deeper} to model the ambiguous mapping between monocular images and depth maps. It can output the depth maps with higher resolution, at the same time reduce the amount of parameters and train on one order of magnitude fewer data. This paper further proposes a scheme for up-convolutions and combine it with the concept of residual learning to create up-projection blocks for the effective up-sampling of feature maps, which has been proved to be more applicable when addressing high-dimensional regression tasks. 

Fully connected layers in some CNN architectures will generate considerable parameters, causing several problems like slower training time, much memory consumption, and chances of overfitting. Some networks in \cite{long2015fully} \cite{zhu2016real} replace fully connected layers with convolutional layers, which can decrease the image matrix to a lower dimension and reduce the amount of parameters. However, convolutional layers are not capable to encode the position and orientation of objects and lack the ability to be spatially invariant to the input data. Thus the output accuracy is much lower than that of fully connected layers. Some further improvements have been proposed to compensate the loss like the convolutional residual networks \cite{laina2016deeper} we introduced above and the dilated convolution we will discuss in next section.

\subsection{Dilated Convolution}
The dilated convolution is a type of convolution that expands the kernel by inserting holes between the kernel elements. The standard convolution operator is defined in \cite{yu2015multi} as:

\begin{equation}
\label{eq:1}
(F * k)(\mathbf{p}) = \sum_{\mathbf{s}+\mathbf{t}=\mathbf{p}}F(\mathbf{s})k(\mathbf{t})
\end{equation}

The dilated convolution operator has been referred to as convolution with a dilated filter. We refer to $l$ as a dilation factor and the l-dilated convolution operator $*_l$ can be defined as:

\begin{equation}
\label{eq:2}
(F *_l k)(\mathbf{p}) = \sum_{\mathbf{s}+l \mathbf{t}=\mathbf{p}}F(\mathbf{s})k(\mathbf{t})
\end{equation}

When $l = 1$, the discrete convolution is simply the 1-dilated convolution. The dilation factor $l$ should be increased exponentially at each layer when building a network with multiple dilated convolution layers. At the same time, the number of parameters associated with each dilated convolution layer is identical. That's the reason why dilated convolution can significantly reduce the amount of parameters. 

Dilated convolution is widely adopted to enhance the performance in computer vision tasks. The multi-scale context architecture with dilated convolutions presented in \cite{yu2015multi} has increased the precision of the advanced semantic segmentation models effectively. \cite{chen2014semantic} adopts dilated convolution in deep convolutional neural networks, which increases the dense computation of neural net responses and achieves a higher accuracy than state-of-the-art algorithms at semantic segmentation task. A network for congested scene recognition (CSRNet) in \cite{li2018csrnet} is easily trained by replacing pooling layers with dilated kernels. And the results demonstrate that CSRNet improves the output performance significantly with lower mean absolute error (MAE) than state-of-the-art algorithms. 

The Alexnet and VGG-16 models apply the pooling layers to downsample the input images and simultaneously extend the receptive field. However, this process generally results in a loss of pixel-wise details. Dilated convolutions can avoid the similar resolution degradation issue but obtain the same computation as pooling layers in Alexnet and VGG-16 networks. Derived from the application of dilated convolutions on semantic segmentation, a similar technique is implemented to tackle the depth estimation task. Dilated convolutions hold the superiority with exponentially expanding of the receptive fields, which supports obtaining the global relationships among pixels in an image without the resolution reduction when decreasing the amount of parameters. 

\section{Dilated Fully CNN Architecture}
\subsection{Overview of the Method}
\begin{figure}[H]
	\includegraphics[width=3.4 in]{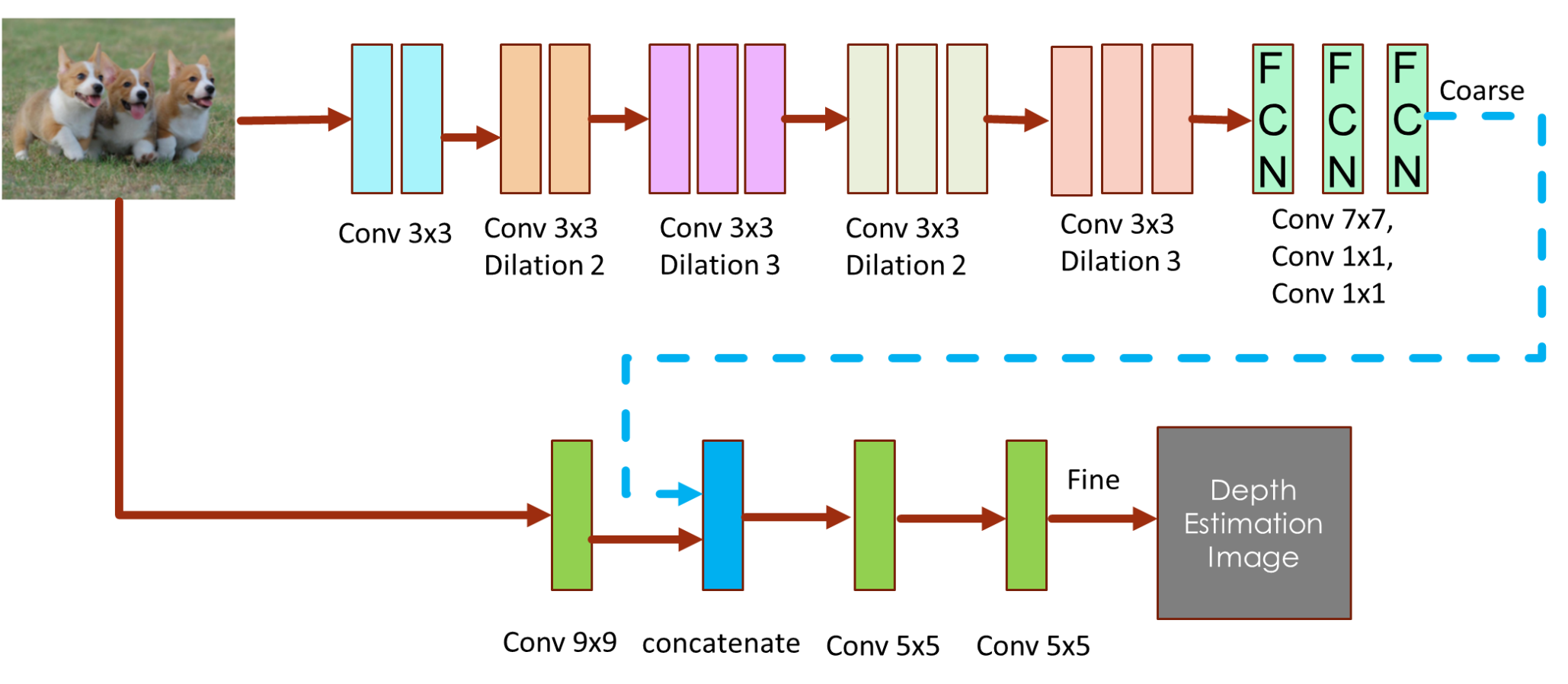}
	\caption{The architecture of two-stages dilated fully convolutional network}
	\label{fig:The architecture of dilated fully convolutional network.}
\end{figure}

Our dilated fully CNN architecture designed for depth estimation is presented in Fig. \ref{fig:The architecture of dilated fully convolutional network.}. It is built upon the multi-scale deep network in \cite{eigen2014depth}, which incorporates two stages: the global coarse-scale network and local fine-scale network.

The upper component (Stack 1) is the global coarse-scale network, which is similar to the VGG-16 network and designed to predict the coarse depth information. The convolutional layers and the fully connected layers in VGG-16 are replaced with the dilated convolutions and convolutional layers respectively. The coarse stage contains dilation layers and fully connected layers (FCN). The dilation layers apply 3x3 convolutions with different dilation factors. The dilations are 1, 2, 3, 2, 3 and 4. Rather than reducing the feature map sizes by convolutional layers and implementing fully connected layers to learn details over the local scene, our network can remain the same feature map sizes with much fewer parameters and simultaneously obtain an overall view of the scene from fully connected layers without the resolution loss. 

The bottom component (Stack 2) is taken as the local fine-scale network. While the coarse stage captures the global scene, we also need to obtain local information in the refined stage. The inputs images will pass a 9x9 convolutional layer with pooling. Then its output and the low-level feature maps output from coarse stage will be concatenated.  

The architecture of the proposed dilated fully CNN with structural parameters is demonstrated in Table \ref{tab:Module architecture}. In the coarse stage of VGG-16 network, the original frames of size 160x120 pixels are down-sampled by each convolutional layer. The front-end module that provides the input to the coarse stage of our network produces feature maps at 80x60 resolution. Our architecture remains the same resolution in the whole framework by replacing the convolutional layers with dilation convolutions. Table \ref{tab:Module architecture} demonstrates that the dilated convolution layers are particularly suited to coarse prediction due to its advantage of expanding the receptive field without the resolution reduction.

\begin{table}[H]
  \centering
  \caption{Module architecture with the image size (size), the convolutional layer number (conv), the channel number (chan), the kernel size, and the dilation layer number in each layer. }
  \scalebox{0.6}{
    \begin{tabular}{c|c|cccccccc|c}
    \hline
          & Layer & 1.1   & 1.2   & 1.3   & 1.4   & 1.5   & 1.6   & 1.7   & 1.8   & upsamp \bigstrut\\
    \hline
    \multicolumn{1}{|c|}{\multirowcell{4}{Stack 1 \\ (VGG)}}
          & size & 160x120 & 80x60 & 40x30 & 20x15 & 10x7  & 1x1   & 1x1   & 1x1   & 80x60\\
          & conv & 2       & 2     & 3     & 3     & 3     & -     & -     & -     & - \\
          & chan  & 64    & 128   & 256   & 512   & 512   & 4096  & 4096  & 4800  & 1 \\
          & ker.sz & 3x3   & 3x3   & 3x3   & 3x3   & 3x3   & -     & -     & -     & - \\
    \hline
    \multicolumn{1}{|c|}{\multirowcell{5}{Stack 1 \\ (OUR)}} 
          & size  & 80x60 & 80x60 & 80x60 & 80x60 & 80x60 & 80x60 & 80x60 & 80x60 & - \\
          & conv  & 2     & 2     & 3     & 3     & 3     & 1     & 1     & 1     & - \\
          & chan  & 64    & 128   & 256   & 512   & 512   & 512   & 512   & 1     & - \\
          & ker.sz & 3x3   & 3x3   & 3x3   & 3x3   & 3x3   & 7x7   & 1x1   & 1x1   & - \\
          & dilation & 1     & 2     & 3     & 2     & 3     & 4     & -     & -     & - \\
    \hline
          & Layer & 2.1   & 2.2   & 2.3   & 2.4   &       &       &       &       & final \\
    \hline
    \multirowcell{4}{Stack 2} & size  & 80x60 & 80x60 & 80x60 & 80x60 &       &       &       &       & 80x60 \\
          & conv  & 1     & -     & 1     & 1     &       &       &       &       & - \\
          & chan  & 63    & -     & 64    & 1     &       &       &       &       & 1 \\
          & ker.sz & 9x9   & -     & 5x5   & 5x5   &       &       &       &       & - \\
    \hline
    \end{tabular}%
  }
  
  \label{tab:Module architecture}%
\end{table}%

\subsection{Training Loss}
Due to the ambiguity of the scale over a global scene in depth estimation, majority of the error is generated when obtaining the average scale of a scene. In order to compensate this training loss, we adopt the scale-invariant mean squared error in \cite{eigen2014depth} as our loss function. Regardless of the absolute global scale, the scale-invariant error is applied to measure correlation among local pixels in the scene, which is defined in as:
\begin{equation} \label{eq:3}
    \textbf{\textit{D}} = \frac{1}{2n}\sum_{i=1}^{n}(\log{y_{i}} - \log{y_{i}^{*}} + \alpha(y,y^{*}))^{2}
\end{equation}
where $y$ and $y^{*}$ represent the estimated depth map and the ground truth respectively. Each of them contains $n$ pixels that are indexed by $i$. And $\alpha(y,y^{*})$ aims to minimize the error between given $y$ and $y^{*}$, which is defined as:
\begin{equation} \label{eq:3.5}
    \alpha(y,y^{*}) = \frac{1}{n}\sum_{i}(\log{y_{i}^{*}} - \log{y_{i}})
\end{equation}
For any predicted depth map $y$, $e^{\alpha}$ is the scale that best aligns it to the ground truth $y^{*}$. The scale-invariant means that the error won't change with any scalar multiples of $y$. 

We refer to this scale-invariant error as our training loss function that is formulated as:
\begin{equation} \label{eq:4}
	\textbf{\textit{L}} = \frac{1}{2\textit{n}^{2}}\sum_{i,j}((\log{y_{i}} - \log{y_{j}}) - (\log{y^{*}_{i}}-\log{y^{*}_{j}}))^{2}
\end{equation}
where $i, j \in {\{1, 2 \ldots n-1\}}$.

The difference between the estimated depth map $y$ and the ground truth $y^{*}$ at pixel $i$ is defined as:
\begin{equation} \label{eq:4.5}
    d_{i} = \log{y_{i}} - \log{y^{*}_{i}}
\end{equation}
Then (\ref{eq:4}) can be re-formulated as: 
\begin{equation} \label{eq:5}
	\textbf{\textit{L}} = \frac{1}{\textit{n}} \sum_{i} \textbf{\textit{d}}^{2}_{i} - \frac{1}{\textit{n}^{2}} \sum_{i,j} \textbf{\textit{d}}_{i}  \textbf{\textit{d}}_{j} 
\end{equation}
	
(\ref{eq:5}) measures the error from the relationships between the output pixel $i$ and pixel $j$. And each pair of predicted pixels and its corresponding ground truth pixels should share a similar amount of the difference between pixels, which can further reduce the errors. 

\section{Experiment Setup}
\subsection{Datasets and Implementation}
We setup the experiment based upon NYU Depth V2 datasets \cite{silberman2012indoor}, which is one of the largest RGB-D image datasets for indoor scene reconstruction. The raw dataset consists of 1449 RGB images with detailed object labels and annotattions with physical relations. Comprising 464 different indoor scenes and classified by 26 scene classes, those images are captured from a variety of buildings in modern cities. NYU Depth V2 dataset is significantly larger and more diverse than another similar Kinect scene dataset - NYU indoor scene dataset, which has limited diversity with only 67 scenes. In order to reduce the chances of overfitting, we shuffle the entire dataset. During the training process, only 800 images of the raw distribution are required. Then we take 200 images to execute the validation test and test our network with 449 images.

\subsection{Training Procedure}

In our network, we are committed to the universality of the structure. Since the image dataset will not generate any additional processing, our proposed network can be simply applied on other datasets. 

Our network is implemented on Keras library running on top of Tensorflow. We train the network in two phases and use the GPU acceleration to speed up the training. First, dilated convolution layers and fully connected layers in the coarse-scale network are pretrained on the NYU Depth V2 training dataset. During this process, the parameters stay identical. Second, the coarse stage outputs concatenated with the edge features of input images are referred as the input images of the fine-scale stage. The convolutional layers in the fine-scale network refine the coarse prediction by aligning it with the detailed information in a local scene.

To make a fair comparison with the performance of traditional VGG-16 framework, two frameworks are both trained on NYU Depth V2 dataset. And they generate output images with the same size, 80x60 for each. The size of input image in VGG-16 framework is 320x240 and the size of input image in our proposed dilated fully convolutional network is 80x60. We refer to stochastic gradient descent (SGD) as the optimizer with the mini-batch size of 16 in the experiment. And we set the learning rate as 0.1 and the momentum as 0.9 for both global coarse-scale stage and local fine-scale stage.

\section{Experiment Results and Analysis}
\subsection{Parameters Comparison}
\begin{figure}[H]
	\subfigure[]
	{
		\includegraphics[width=3.2in]{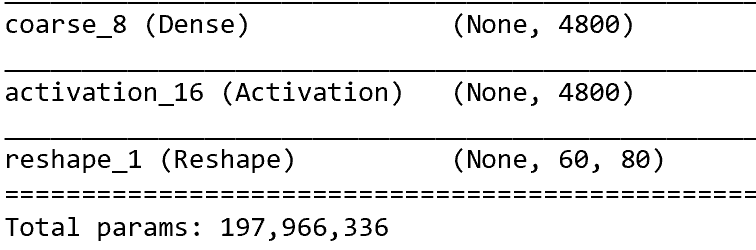}
		\label{b_1}
	}
	\subfigure[]
	{
		\includegraphics[width=3.2in]{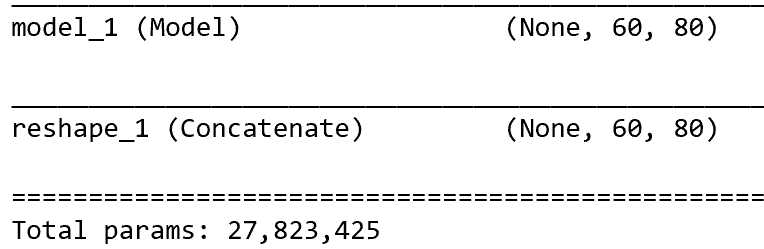}
		\label{b_2}
	}
	\subfigure[]
	{
		\includegraphics[width=3.2in]{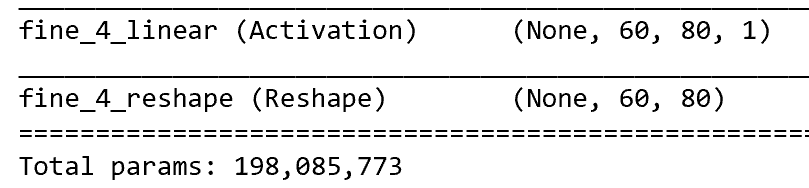}
		\label{b_3}
	}
	\subfigure[]
	{
		\includegraphics[width=3.2in]{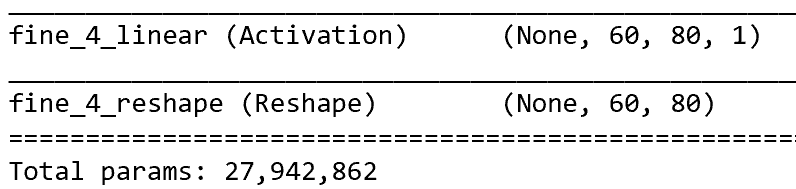}
		\label{b_4}
	}
	\caption{Comparison on the scale of parameters generated in CNN. (a) and (c) present the number of total parameters in the coarse-scale stage (Stack 1) and the whole framework based on VGG-16 network, (b) and (d) present the number of total parameters in the coarse-scale stage (Stack 1) and the whole framework based on our proposed network.}
	\label{fig:The number of parameters of two architecture}
\end{figure}

Fig. \ref{fig:The number of parameters of two architecture} demonstrates the effectiveness of dilated convolutions on decreasing the amount of parameters. "Total params" in (a) and (c) represents the amount of total parameters of the coarse-scale stage based on VGG-16 network and our proposed network respectively. (a) and (d) present the total parameters in the whole framework of VGG-16 network and our proposed architecture. Since the coarse-scale network is more complex with more layers than the fine-scale network, majority of parameters are generated from the coarse-scale stage, which can be observed from the comparison between total parameters in (a) and (c).

The experimental results demonstrate that our proposed network achieves over seven times reduction of the parameters amount both in the coarse-scale stage and the whole framework compared with the conventional VGG-16 network. Thus our proposed network can liberate considerable computation resources during the training process, which benefits from the significant parameters reduction. Besides, the limited scale of parameters makes the proposed network a viable candidate to be applied to other embedded architectures. 

\subsection{Depth Estimation Results}
\begin{figure}[H]
    \flushleft
    \scalebox{0.75}
{
	\subfigure[]
	{
	    \begin{minipage}[t]{0.25\linewidth}
		\includegraphics{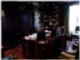}\vspace{4pt}			\includegraphics{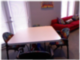}\vspace{4pt}
		\includegraphics{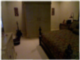}\vspace{4pt}
		\includegraphics{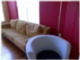}\vspace{4pt}
		\includegraphics{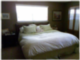}\vspace{4pt}
		\includegraphics{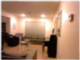}\vspace{4pt}
		\includegraphics{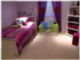}\vspace{4pt}
		\includegraphics{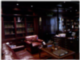}\vspace{4pt}
		\includegraphics{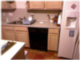}\vspace{4pt}
		\includegraphics{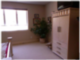}\vspace{4pt}
		\includegraphics{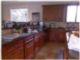}\vspace{4pt}
		\includegraphics{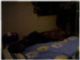}\vspace{4pt}
		\end{minipage}
	}
	\hspace{4pt}
	\subfigure[]
	{
		\begin{minipage}[t]{0.25\linewidth}
		\includegraphics{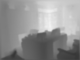}\vspace{4pt}	\includegraphics{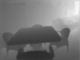}\vspace{4pt}
		\includegraphics{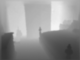}\vspace{4pt}
		\includegraphics{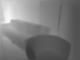}\vspace{4pt}
		\includegraphics{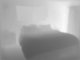}\vspace{4pt}
		\includegraphics{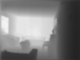}\vspace{4pt}
		\includegraphics{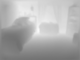}\vspace{4pt}
		\includegraphics{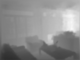}\vspace{4pt}
		\includegraphics{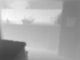}\vspace{4pt}
		\includegraphics{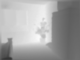}\vspace{4pt}
		\includegraphics{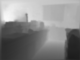}\vspace{4pt}
		\includegraphics{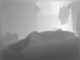}\vspace{4pt}
		\end{minipage}
	}
	\hspace{4pt}
	\subfigure[]
	{
		\begin{minipage}[t]{0.25\linewidth}
		\includegraphics{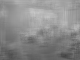}\vspace{4pt}
		\includegraphics{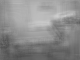}\vspace{4pt}
		\includegraphics{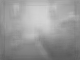}\vspace{4pt}
		\includegraphics{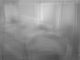}\vspace{4pt}
		\includegraphics{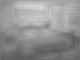}\vspace{4pt}
		\includegraphics{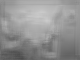}\vspace{4pt}
		\includegraphics{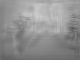}\vspace{4pt}
		\includegraphics{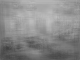}\vspace{4pt}
		\includegraphics{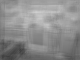}\vspace{4pt}
		\includegraphics{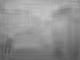}\vspace{4pt}
		\includegraphics{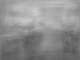}\vspace{4pt}
		\includegraphics{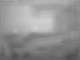}\vspace{4pt}
		\end{minipage}
	}
	\hspace{4pt}
	\subfigure[]
	{
		\begin{minipage}[t]{0.25\linewidth}
		\includegraphics{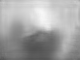}\vspace{4pt}
		\includegraphics{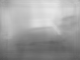}\vspace{4pt}
		\includegraphics{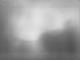}\vspace{4pt}
		\includegraphics{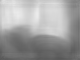}\vspace{4pt}
		\includegraphics{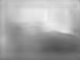}\vspace{4pt}
		\includegraphics{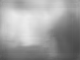}\vspace{4pt}
		\includegraphics{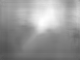}\vspace{4pt}
		\includegraphics{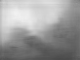}\vspace{4pt}
		\includegraphics{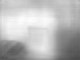}\vspace{4pt}
		\includegraphics{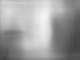}\vspace{4pt}
		\includegraphics{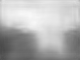}\vspace{4pt}
		\includegraphics{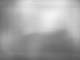}\vspace{4pt}
		\end{minipage}
	}
}

	\caption{Example depth estimations. (a) and (b) represent the input images and the ground truth images respectively, (c) refers to the output depth estimation from VGG-16 network, (d) refers to the output depth estimation from our proposed network }
	\label{fig:Comparison on depth estimation}
\end{figure}

We propose a dilated fully convolutional neural network which is designed upon the architecture in \cite{eigen2014depth}, and evaluate its depth estimation performance compared with the conventional VGG-16 network. The experimental results are presented in Fig. \ref{fig:Comparison on depth estimation}.

The comparison between output images and ground truth images demonstrates that the dilated convolutions in our architecture can still obtain the related details between local pixels in a patch and the global view of a scene even without the fully connected layers. 

Since fully connected layers learn all combinations of the features of the previous layers, the VGG-16 module achieves a good performance in some local spatial details. Comparing the predicted depth information based on VGG-16 module and our network, it is observed that our architecture can't perform as good as VGG-16 module when predicting depth maps around the object contours. The ambiguous pixels are generated during the training process, where majority of the output depth information will be blurred by the missing pixels around objects boundaries and reflective surfaces. Several techniques have been proposed to predict the boundary type of the superpixel edges, such as the discrete-continuous depth estimation approach in \cite{liu2014discrete}. However, our model focuses on the enhancement over a global view of the scene and ignores the boundary performance. It's obviously observed that our network outperforms traditional VGG-16 network a lot globally, with much higher resolutions and more realistic representation of the relationships between objects and the environment. 

\section{Conclusion}

In this paper, we proposed a dilated fully convolutional neural network to predict the depth information from a single image. The network is designed based on the multi-scale deep network in \cite{eigen2014depth}, which contains two stages: the coarse-scale network and the fine-scale network. The coarse stage predicts the global depth information, which is refined locally in the fine-tune stage. We replace the convolutional layers and fully connected layers in the coarse stage with dilated convolutions and convolutional layers respectively. By implementing the dilated convolutions, the network can reduce the amount of parameters significantly without the resolution reduction, which benefits from the exponential expansion of receptive fields in dilated convolutions.
And the experiment results demonstrate that our proposed network achieves the state-of-the-art performance on depth estimation for NYU Depth V2 datasets. The output depth information outperforms VGG-16 network with higher resolutions and more realistic representation of the relationships between objects and the environment, while generating much fewer parameters and releasing more memory resources.  

Predicting the depth information around the objects boundaries is a weak point of our proposed network. In future work, we will incorporate specific techniques into our network and effectively enhance the depth prediction performance around objects boundaries. Besides, our network only evaluates the visual performance on a single dataset, which is insufficient to demonstrate all the superiority. We plan to further evaluate the predicted depth information on several effective image criteria, which can represent the strong points that are not presented visually. And we will apply the dilated convolutions to more advanced CNNs to further improve the depth prediction performance.

\footnotesize{


\bibliographystyle{refastesj}
\bibliography{astesj.bib}

} 

\end{multicols}

\end{document}